\documentclass[10pt,twocolumn,letterpaper]{article}

\usepackage[accsupp]{axessibility}
\usepackage[pagenumbers]{wacv} 

\usepackage{graphicx}
\usepackage{amsmath, amsthm, amssymb, amsfonts}
\usepackage{booktabs}
\usepackage{makecell}
\usepackage{comment}
\usepackage{multirow}
\usepackage{color, colortbl}
\usepackage[table]{xcolor}
\definecolor{LightCyan}{rgb}{0.88,1,1}

\usepackage[colorlinks = true,
            linkcolor = purple,
            urlcolor  = blue,
            citecolor = cyan,
            anchorcolor = black]{hyperref}

\usepackage{textcomp}

\usepackage[capitalize]{cleveref}
\crefname{section}{Sec.}{Secs.}
\Crefname{section}{Section}{Sections}
\Crefname{table}{Table}{Tables}
\crefname{table}{Tab.}{Tabs.}

\usepackage{xr}
\makeatletter
\begin{document}

\title{HaGRIDv2: 1M Images for Static and Dynamic Hand Gesture Recognition}

\author{
Anton Nuzhdin \\
{\tt\small anton.nuzhdin.hse@gmail.com}
\and
Alexander Nagaev \\
{\tt\small sashanagaev1111@gmail.com}
\and
Alexander Sautin \\
{\tt\small sasautin007@gmail.com}
\and
Alexander Kapitanov \\
{\tt\small kapitanovalexander@gmail.com}
\and
Karina Kvanchiani \\
{\tt\small karinakvanciani@gmail.com}
\\
\\
\hspace{-12em}SberDevices, Russia
}

\maketitle
\begin{abstract}
This paper proposes the second version of the widespread Hand Gesture Recognition dataset HaGRID -- HaGRIDv2. We cover 15 new gestures with conversation and control functions, including two-handed ones. Building on the foundational concepts proposed by HaGRID's authors, we implemented the dynamic gesture recognition algorithm and further enhanced it by adding three new groups of manipulation gestures. The ``no gesture" class was diversified by adding samples of natural hand movements, which allowed us to minimize false positives by 6 times. Combining extra samples with HaGRID, the received version outperforms the original in pre-training models for gesture-related tasks. Besides, we achieved the best generalization ability among gesture and hand detection datasets. In addition, the second version enhances the quality of the gestures generated by the diffusion model. HaGRIDv2, pre-trained models, and a dynamic gesture recognition algorithm are publicly available.
\end{abstract}

\section{Introduction}
\label{sec:intro}

Hand gestures, as natural and intuitive expressions, effectively reflect emotions and facilitate communication~\cite{general_gesture}. Their ability to convey messages quickly makes gestures invaluable for Human-Computer Interaction (HCI)~\cite{hci}. Thus, the development of Hand Gesture Recognition (HGR) systems has the potential to significantly enhance user interfaces in various domains~\cite{gesture_review, gesture_applications} such as robotic control~\cite{robot_control}, driver assistance~\cite{gesture_cars}, and medicine~\cite{mems, medicine, medicine2} for touchless interaction. The proposed research aims to develop a comprehensive HGR system for video conferencing~\cite{zoom_ges_rec, google_ges_rec} and home automation devices~\cite{home_automation, home_automation3, device_control, home_automation1, home_automation2}. The system should enhance participants' communication, enable remote control of device functions~\cite{zoomtouch, device_control2}, allow the manipulation of various objects on the screen, and activate different platform features~\cite{features_control}. Considering the described application area, the system should enable intuitive operation through easy-to-demonstrate functional gestures, offer instant feedback, and efficiently operate on resource-constrained edge devices.

\begin{figure}[t]
  \centering
  \includegraphics[width=0.95\linewidth]{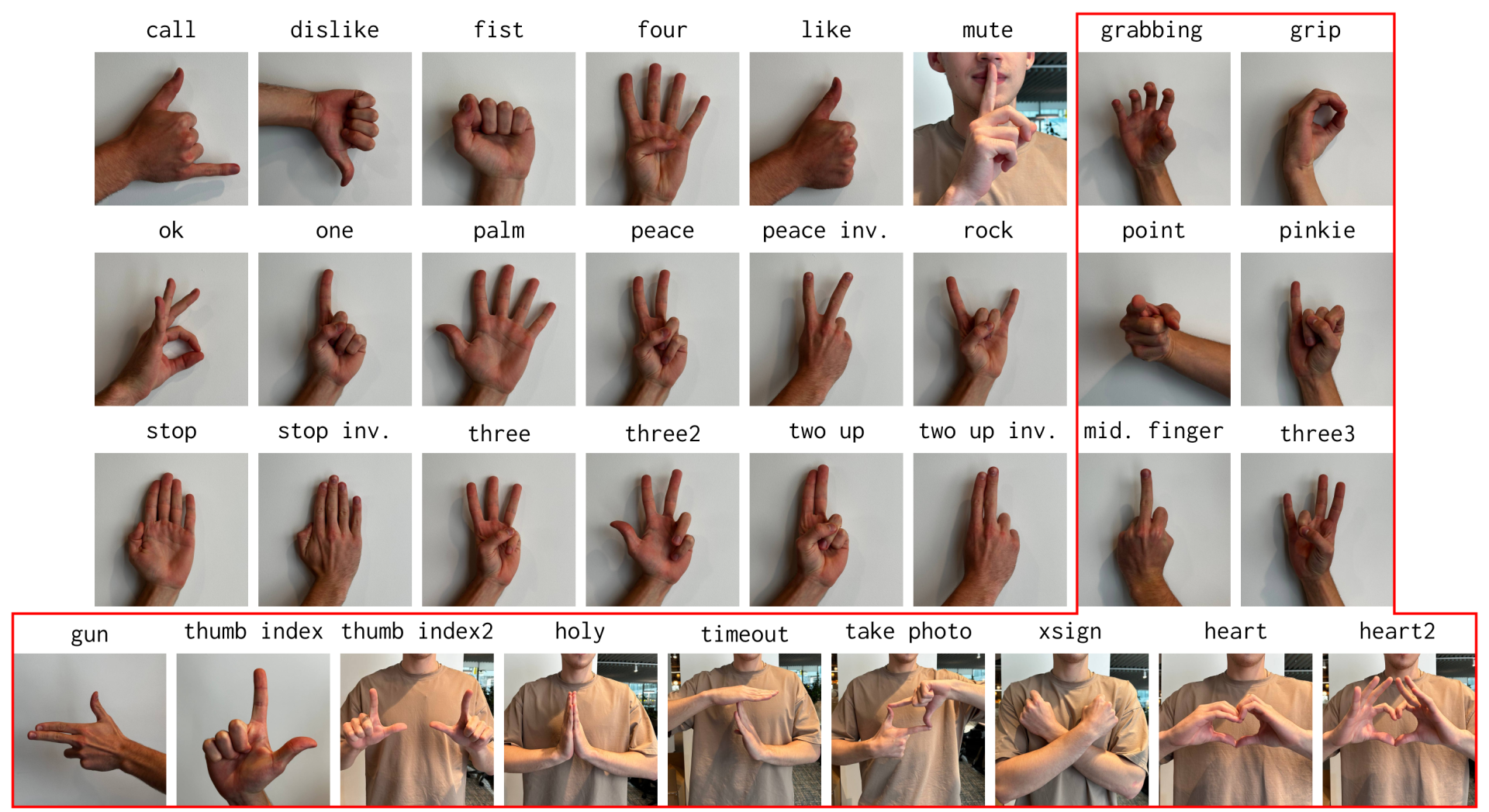}
  \caption{The 15 outlined in red new gesture classes added to HaGRID's 18 ones (``inv" stands for ``inverted").}
  \label{fig: gestures}
\end{figure}

Neural networks have recently become the primary component of HGR systems~\cite{ges_rec, nvgesture, ges_rec2} and action recognition in general~\cite{tsm, c3d, i3d}. Large datasets aligned with system constraints are required to train a model resilient to real-world conditions. Based on the requirements described above, the appropriate dataset should include a range of gestures categorized as manipulative (e.g., clicking, swiping, or zooming the screen), control (e.g., taking screenshots), and conversational (e.g., expressing approval, disapproval, love, anger, regret)~\cite{taxomony, taxomony2}. The first category usually implies dynamic gestures, while the last two involve static ones. Thus, the dataset should encompass static and dynamic gestures (i.e., videos rather than images) to cover all three categories effectively. The HGR model should efficiently provide real-time inference on the CPU while remaining lightweight, as it should be integrated into the resource-constrained device. Compliance with such requirements significantly restricts the choice of neural network architectures capable of operating in the temporal dimension.

Most existing gesture datasets are limited in several senses: some overlook functional and commonsense gestures, and others only cover static or dynamic gestures. The most suitable dataset, HaGRID~\cite{hagrid}, includes various conversational and control gestures and proposes an algorithm for dynamic manipulative gesture recognition. The algorithm required static gestures to construct dynamic ones, satisfying the conditions of the lightweight model. The authors~\cite{hagrid} pursued the goal of creating the HGR system for home automation and video conferencing services. Therefore, all chosen gestures are straightforward and perform a specific function. However, HaGRID lacks support for essential device interaction gestures, such as clicking, zooming, taking screenshots, and cursor control. Besides, the extra ``no gesture" class contains too homogeneous hands, provoking the model to predict false positives. In this regard, we decided to expand the dataset with new gestures to support the development of dynamic gestures, introduce a new ``no gesture" class, and enhance the dynamic gesture recognition algorithm.

\begin{table*}[]
\centering
\scalebox{0.75}{
\resizebox{\textwidth}{!}{%
\begin{tabular}{llllllll}
\hline
Dataset & \multicolumn{1}{c}{Samples} & \multicolumn{1}{c}{Classes} & \multicolumn{1}{c}{Subjects} & \multicolumn{1}{c}{Scenes} & \multicolumn{1}{c}{Resolution} & \multicolumn{1}{c}{Annotations} & \multicolumn{1}{c}{Annotation Method} \\ \hline\hline
\multicolumn{1}{c}{} & \multicolumn{1}{c}{} & \multicolumn{1}{c}{} & \multicolumn{1}{c}{} & \multicolumn{1}{c}{Gesture Detection} & \multicolumn{1}{c}{} & \multicolumn{1}{c}{} \\ \hline
LaRED, 2014~\cite{lared} & \multicolumn{1}{c}{243,000} & \multicolumn{1}{c}{81} & \multicolumn{1}{c}{10}  & \multicolumn{1}{c}{10}   & \multicolumn{1}{c}{640$\times$480}  & \multicolumn{1}{c}{masks} & \multicolumn{1}{c}{automatically} \\
OUHANDS, 2016~\cite{ouhands} & \multicolumn{1}{c}{3,000}& \multicolumn{1}{c}{10} & \multicolumn{1}{c}{23} & \multicolumn{1}{c}{various} & \multicolumn{1}{c}{640$\times$480}  & \multicolumn{1}{c}{masks, boxes} & \multicolumn{1}{c}{automatically} \\
HANDS, 2021~\cite{hands} & \multicolumn{1}{c}{12,000}  & \multicolumn{1}{c}{29} & \multicolumn{1}{c}{5} & \multicolumn{1}{c}{5} & \multicolumn{1}{c}{960$\times$540}  & \multicolumn{1}{c}{boxes} & \multicolumn{1}{c}{-} \\
SHAPE, 2022~\cite{shape} & \multicolumn{1}{c}{33,471}  & \multicolumn{1}{c}{31+1} & \multicolumn{1}{c}{20} & \multicolumn{1}{c}{various} & \multicolumn{1}{c}{4128$\times$3096} & \multicolumn{1}{c}{masks, boxes} & \multicolumn{1}{c}{manually} \\
HaGRID, 2023~\cite{hagrid} & \multicolumn{1}{c}{554,800} & \multicolumn{1}{c}{18+1} & \multicolumn{1}{c}{37,583}& \multicolumn{1}{c}{$\ge$37,583} & \multicolumn{1}{c}{1920$\times$1080} & \multicolumn{1}{c}{boxes, keypoints*} & \multicolumn{1}{c}{manually} \\
\rowcolor{LightCyan}
HaGRIDv2, 2024 (ours) & \multicolumn{1}{c}{1,086,158} & \multicolumn{1}{c}{33+1} & \multicolumn{1}{c}{65,977} & \multicolumn{1}{c}{$\ge$65,977} & \multicolumn{1}{c}{1920$\times$1080} & \multicolumn{1}{c}{boxes, keypoints*} & \multicolumn{1}{c}{mixed}\\

\hline
\multicolumn{1}{c}{} & \multicolumn{1}{c}{} & \multicolumn{1}{c}{} & \multicolumn{1}{c}{} & \multicolumn{1}{c}{Gesture Classification} & \multicolumn{1}{c}{} & \multicolumn{1}{c}{} \\ \hline
HTU HGR, 2011~\cite{ntuhgr}  & \multicolumn{1}{c}{1,000} & \multicolumn{1}{c}{10}& \multicolumn{1}{c}{10} & \multicolumn{1}{c}{various} & \multicolumn{1}{c}{640$\times$480} & \multicolumn{1}{c}{labels} & \multicolumn{1}{c}{manually} \\
Kinect Leap, 2014~\cite{kinect_leap1, kinect_leap2} & \multicolumn{1}{c}{1,400} & \multicolumn{1}{c}{10} & \multicolumn{1}{c}{14} & \multicolumn{1}{c}{-} & \multicolumn{1}{c}{640$\times$480}  & \multicolumn{1}{c}{labels} & \multicolumn{1}{c}{manually} \\
Senz3D, 2015~\cite{senz3d1, senz3d2} & \multicolumn{1}{c}{1,320} & \multicolumn{1}{c}{11} & \multicolumn{1}{c}{4} & \multicolumn{1}{c}{-} & \multicolumn{1}{c}{640$\times$480} & \multicolumn{1}{c}{labels} & \multicolumn{1}{c}{manually} \\
SIT-HANDS, 2023~\cite{sithands} & \multicolumn{1}{c}{4,200} & \multicolumn{1}{c}{14} & \multicolumn{1}{c}{10} & \multicolumn{1}{c}{various} & \multicolumn{1}{c}{1920$\times$1080} & \multicolumn{1}{c}{labels} & \multicolumn{1}{c}{manually} \\

\hline
\multicolumn{1}{c}{} & \multicolumn{1}{c}{} & \multicolumn{1}{c}{} & \multicolumn{1}{c}{} & \multicolumn{1}{c}{Hand Detection} & \multicolumn{1}{c}{} & \multicolumn{1}{c}{} \\ \hline
HAND, 2011~\cite{hand} & \multicolumn{1}{c}{5,628} & \multicolumn{1}{c}{1} & \multicolumn{1}{c}{-} & \multicolumn{1}{c}{-} & \multicolumn{1}{c}{mixed} & \multicolumn{1}{c}{boxes}  & \multicolumn{1}{c}{automatically} \\
EgoHands, 2015~\cite{egohands} & \multicolumn{1}{c}{4,800} & \multicolumn{1}{c}{1} & \multicolumn{1}{c}{various} & \multicolumn{1}{c}{various} & \multicolumn{1}{c}{720$\times$1280}  & \multicolumn{1}{c}{masks, boxes} & \multicolumn{1}{c}{mixed}\\
Human-Parts, 2019~\cite{human_parts} & \multicolumn{1}{c}{14,962}  & \multicolumn{1}{c}{3} & \multicolumn{1}{c}{-} & \multicolumn{1}{c}{various} & \multicolumn{1}{c}{mixed} & \multicolumn{1}{c}{boxes} & \multicolumn{1}{c}{automatically} \\
TV-Hands, 2019~\cite{tvhands} & \multicolumn{1}{c}{9,498}& \multicolumn{1}{c}{1} & \multicolumn{1}{c}{various} & \multicolumn{1}{c}{various} & \multicolumn{1}{c}{-} & \multicolumn{1}{c}{boxes} & \multicolumn{1}{c}{manually} \\
ContactHands, 2020~\cite{contacthands} & \multicolumn{1}{c}{20,516}& \multicolumn{1}{c}{1} & \multicolumn{1}{c}{-} & \multicolumn{1}{c}{various} & \multicolumn{1}{c}{-} & \multicolumn{1}{c}{boxes} & \multicolumn{1}{c}{manually} \\
BodyHands, 2022~\cite{bodyhands} & \multicolumn{1}{c}{20,490}  & \multicolumn{1}{c}{2} & \multicolumn{1}{c}{-} & \multicolumn{1}{c}{various} & \multicolumn{1}{c}{-} & \multicolumn{1}{c}{boxes} & \multicolumn{1}{c}{manually} \\
\hline\hline

\hline
\end{tabular}}}

\caption{The main parameters of the most popular gesture datasets. ``+1" in the third column means the dataset contains an extra class ``no gesture”. ``-" in some columns means information was not found. * — keypoints prepared by using the MediaPipe~\cite{mediapipe} hand model.}
\label{tab:datasets_table}
\end{table*}

It was decided to expand the HaGRID dataset by adding new classes. This paper introduces HaGRIDv2, the second version of the HaGRID dataset, designed to enrich the functionality of HGR systems for video conferencing and home automation. The contribution of this paper is three-fold:

\begin{itemize}
    \item HaGRIDv2 incorporates 15 new gesture classes (\cref{fig: gestures}) performing control and conversational functions. The cross-dataset evaluation experiments confirmed HaGRIDv2's best domain generalization ability among gesture detection datasets (see \cref{tabl:cross_gesture_det}).
    \item The ``no gesture" class is upgraded relative to HaGRID's by incorporating domain-specific natural hand positions allowed to minimize false positives by 6 times (see \cref{fig: false}).
    \item We extend dynamic gesture recognition algorithm~\cite{hagrid} capabilities, developing swipes, clicks, zooms, drag-and-drops, and other manipulative gestures (see \cref{fig: dynamic_ges} in the suppl. materials). 
\end{itemize}

The dataset and the extended algorithm for recognizing dynamic gestures are publicly available\footnote {\url{https://github.com/hukenovs/hagrid}}\footnote{\url{https://github.com/ai-forever/dynamic_gestures}} under the modified Creative Commons CC-BY 4.0 license.

HaGRIDv2 was designed to solve the gesture detection task. However, the proposed paper also inspects the applicability of HaGRIDv2 to address other gesture-related tasks: gesture full-frame classification, hand detection, and text-to-image gesture generation (see \cref{fig: tasks} in the suppl. materials for the difference between the tasks). Although HaGRIDv2 does not focus on hand detection, it performs well on the standard benchmarks (\cref{tabl:cross_hand_det}). \cref{fig: pretrains_classification} shows that the gesture bucket expansion improved the HaGRID pre-training abilities in the gesture tasks. Also, \cref{fig: sbs2} in the suppl. materials demonstrates the HaGRIDv2's capabilities to address the issue of diffusion models producing anatomically incorrect gestures.

\section{Related Work}
\label{sec:related_work}

There are a variety of HGR datasets with gestures categorized based on their applications~\cite{taxomony, taxomony2}: sign language~\cite{asl_dataset}, control~\cite{egogesture, natops, ipn, lared, hagrid, hands, chalean}, conversational~\cite{sithands, hagrid, shape, hands, ouhands, senz3d1, kinect_leap1, lared} and manipulative gestures~\cite{nvgesture, egogesture, jester, ipn, skig, chairgest, dvs128}. The proposed research aims to develop an HGR system for device control and video conferencing, where manipulative, control, and conversational gestures are essential. Therefore, the system should recognize both static and dynamic gestures, which are reviewed in this study. Sign language recognition datasets are excluded because their gestures are unsuitable for performing the described functions.

Dynamic gesture datasets are typically annotated for action recognition, classifying entire video sequences. In contrast, static gesture recognition can be achieved by solving various tasks, including gesture detection and classification (see \cref{fig: tasks} in the suppl. materials for difference), hand gesture keypoint estimation, and gesture segmentation. However, classification labels are impractical in multi-person frames, keypoints can stick together when the person is far away, and segmentation masks are excessive and costly. Additionally, only third-person data are suitable for video conferences and device control.

The overview of the related HGR datasets is organized as follows. \cref{subsec:devices_control} discusses datasets relevant to the device control task, while \cref{subsec:other_tasks} examines datasets related to gestures data domains not central to this research.

\subsection{Devices Control}
\label{subsec:devices_control}

\textbf{Static Gestures}. \cref{tab:datasets_table} shows that only HaGRID~\cite{hagrid}, LaRED~\cite{lared}, OUHANDS~\cite{ouhands}, HANDS~\cite{hands}, and SHAPE~\cite{shape} datasets are relevant for the required annotations. Each dataset has limitations that can affect its suitability for developing a reliable HGR system for real-world conditions. The LaRED~\cite{lared} and the OUHANDS~\cite{ouhands} datasets include images captured from close distances, making them unsuitable for training models operating over larger spaces. Besides, there is no access to the LaRED samples due to the outdated link. Also, OUHANDS~\cite{ouhands} and HANDS~\cite{hands}, constructed with only 23 and 5 subjects, respectively, are inappropriate for developing a robust HGR system. In addition, even the most diverse datasets on classes lack gesture variety for control and conversational purposes. So, while the SHAPE~\cite{shape} dataset lacks enough control gestures, the HaGRID~\cite{hagrid} suffers from a deficiency of conversational ones, containing only ``like" and ``dislike" emotional gestures. Since emotional gestures are essential for video meetings, such an omission restricts the overall functionality of the HGR system. 

\textbf{Dynamic Gestures}. Datasets such as ~\cite{natops, skig, chairgest, chalean, nvgesture, dvs128, egogesture, jester, ipn} are the most relevant for dynamic gesture recognition in device interaction and video conferencing. There are only ChAirGest~\cite{chairgest}, Jester~\cite{jester}, and IPN Hand~\cite{ipn} datasets meeting the described requirements about the existence of functional gestures and third-person view. However, these datasets focus mainly on manipulative dynamic gestures and lack the necessary static conversational and control gestures. This gap highlights their insufficiency in covering the full spectrum of gestures needed for device control and video conferencing.

Although there is no entirely suitable dataset, the HaGRID dataset encompasses control and conversational gestures while simultaneously allowing the recognition of dynamic manipulative ones utilized proposed in \cite{hagrid} algorithm. Therefore, we decided to make some changes and create a second version of HaGRID, adding new gestures and diversifying the ``no gesture" class.

\subsection{Other Tasks}
\label{subsec:other_tasks}
HaGRIDv2's design, with one gesture per frame, supports full-frame classification, which is ideal for single-user interactions with personal devices. Additionally, HaGRIDv2 includes a wide variety of hand postures, including complex ones and natural hand positions, making it well-suited for hand detection tasks.

\textbf{Hand Gesture Classification}. All the static datasets for object detection reviewed below can also be used for image classification. In addition to them, there are NTU HGR~\cite{ntuhgr}, Kinect Leap~\cite{kinect_leap1}, Senz3D~\cite{senz3d1}, and SIT-HANDS~\cite{sithands} (see \cref{tab:datasets_table}). However, these datasets also have disadvantages in solving the human-computer interaction problem. The NTU HGR~\cite{ntuhgr}, Senz3D~\cite{senz3d1}, and Kinect Leap~\cite{kinect_leap1} datasets contain 100-140 samples per gesture class, which is not enough to build a sufficiently high-quality model (see ablation study in \cite{hagrid}). In addition, Senz3D and Kinect Leap scenes are homogeneous, and the gestures are too close to the camera. The SIT-HANDS~\cite{sithands} dataset was made heterogeneous in subjects, lighting conditions, and background. However, the dataset has only 2,800 training frames and contains a relatively small variety of gestures, severely limiting the system's functionality.

\begin{figure*}[tb]
  \centering
  \includegraphics[width=1.0\linewidth]{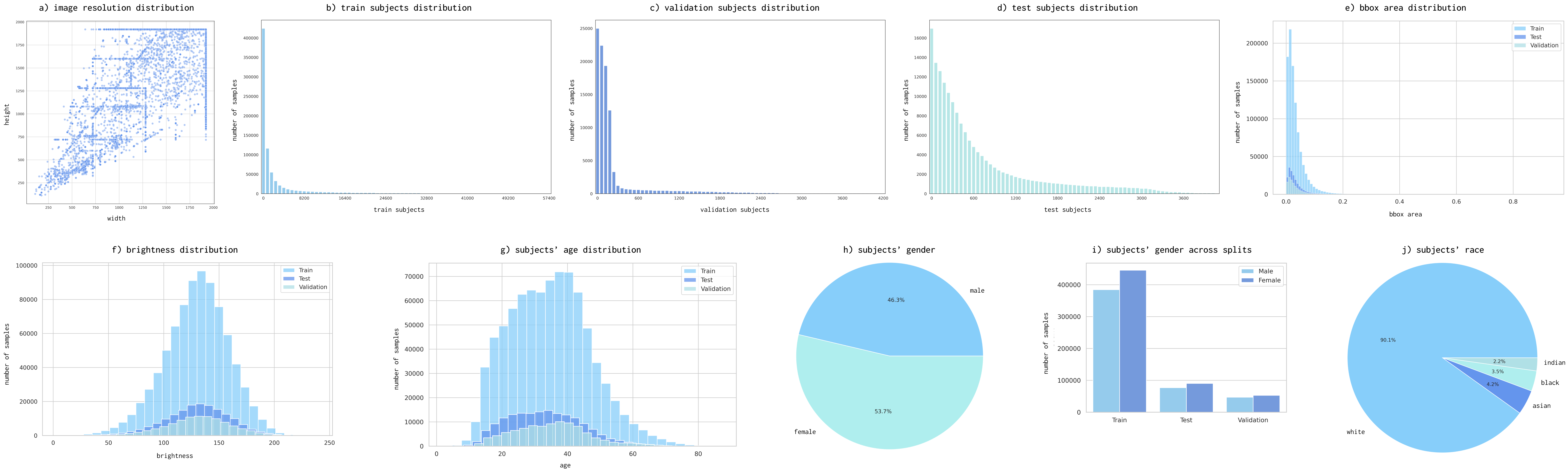}
  \caption{The key statistics of HaGRIDv2. (a) Image resolution distribution showing the scatter of image dimensions; (b-d) Distribution of subjects in the training, validation, and test sets, respectively; (e) Bounding box area distribution; (f) Brightness distribution; (g-i) Age and gender distributions of subjects, received automatically by MiVOLO~\cite{mivolo} neural network; (j) Racial distribution of subjects, received automatically by FairFace~\cite{fairface} neural network.}
  \label{fig: chars}
\end{figure*}

\textbf{Hand Detection}. Among the hand detection datasets there are HAND~\cite{hand}, EgoHands~\cite{egohands}, Human-Parts~\cite{human_parts}, TV-Hands~\cite{tvhands}, ContactHands~\cite{contacthands}, BodyHands~\cite{bodyhands}. Hands detection datasets may differ in the annotation type: some use horizontal aligned bounding boxes, while others use oriented bounding boxes. Since most well-known detection models work with horizontally aligned bounding boxes, this type is preferred, making HAND~\cite{hand}, TV-Hands~\cite{tvhands}, and ContactHands~\cite{contacthands} unsuitable.

Datasets vary in the number of people, subjects, resolution, and scenes. EgoHands~\cite{egohands} is intended for first-person gesture detection and segmentation across various backgrounds, while BodyHands~\cite{bodyhands} and Human-Parts~\cite{human_parts} offer 14,000 and 15,000 diverse third-person samples, respectively. Due to their heterogeneity, these datasets are valuable benchmarks for hand detection, but they focus only on natural hand postures and do not include specific gestures or complex finger positions. By addressing some key limitations in existing datasets, HaGRIDv2 offers improved support for these areas.


\section{HaGRIDv2 Dataset}
\label{sec:dataset}
The HaGRIDv2 dataset differs from the HaGRID with the following key updates: 
\begin{itemize}
    \item We added ``holy", ``heart" (in two variations, see \cref{fig: gestures}), ``middle finger", and ``gun" as emotional conversational gestures used during the conversation, and ``three3" for extra functions.
    \item We expanded the range of control gestures by one-handed ``thumb index", ``grip", ``point", ``pinkie", and ``grabbing", and two-handed ``thumb index2", ``timeout", ``take photo", ``xsign".
    \item Some of added static gesture classes were designed to enable the extension of the dynamic gesture recognition algorithm by developing such gestures as ``drag and drop", ``click", ``zoom in", ``zoom out" and new variations of swipes.
    
    In total, 15 new static gesture classes and four groups of dynamic gestures were developed to cover various device functions (see \cref{fig: gestures} and \cref{fig: dynamic_ges} in the suppl. materials).
    \item The HaGRIDv2's ``no gesture" class includes a broader range of natural hand positions (e.g., relaxed hands near the face, holding a cup, natural gesticulation), whereas HaGRID is limited to a single relaxed hanging hand posture (see \cref{fig: no_gesture_samples} in the suppl. materials). Such natural movements were specifically selected as the most commonly used gestures during video conference meetings.
\end{itemize}

\subsection{Dataset Creating Pipeline}
The data creation pipeline almost followed the one proposed by the original HaGRID authors~\cite{hagrid} to maintain the consistency of HaGRIDv2's data distribution. Also, such a pipeline allows us to collect heterogeneous samples in large volumes. We used the same crowdsourcing platforms such as Yandex.Toloka\footnote{\url{https://toloka.yandex.ru/}} and ABC Elementary\footnote{\url{https://elementary.activebc.ru/}} and instructions for crowdworkers through mining, validation, and filtration steps, described in Dataset Creating Pipeline in \cite{hagrid}. In the mining stage, crowdworkers capture photos with a specified gesture under controlled conditions. In the validation stage, images are reviewed to ensure they meet the criteria, and only correctly executed photos are retained. Finally, the filtration stage aimed to remove ethically sensitive images, including those featuring individuals under 18.

\textbf{Annotation}.
We have decided to replace HaGRID's manual box annotation with an automated one due to its time- and labor-intensive nature. A substantial size of HaGRID is enough to train a robust hand detector for automated annotations. We have also implemented crowd moderation for quality assurance.

The annotation process was divided into hand detection and gesture labeling. We trained the YOLOv10x~\cite{yolov10} detector on the HaGRID, previously reducing all gesture classes to one class ``hand". For images featuring one-handed gestures, the hanging hand is identified as the one that is always below the hand with the gesture (see \cref{fig: auto}a in the suppl. materials). For two-handed gestures, we obtain joint boxes by combining two hand boxes, making a box from the upper left edge of the left hand to the lower edge of the right hand (see \cref{fig: auto}b in the suppl. materials). \cref{fig: auto}c in the suppl. materials shows the exception of the ``xsign" gesture, where two boxes are merged to a square, the side of which is equal to the distance between the extreme points of the boxes.

\subsection{Dataset Characteristics}
\label{subsec:dataset_char}

The HaGRIDv2 dataset is an extension of the widely used HGR dataset HaGRID. Adding 531,358 samples divided into 15 new gesture classes and the extra ``no gesture" class to the original HaGRID, the received combination contains over a million primarily Full HD RGB images (see \cref{fig: gestures} and \cref{fig: chars}d). Since the presented research aims to build a system for home automation devices and conference control, added classes are related to these domains. Thus, each gesture is intended to perform a specific associative function, as shown in \cref{tabl:funcs} in the suppl. materials. A special ``no gesture" class encompasses domain-specific hand postures common for mentioned applications, such as hands near the face, relaxed, or holding objects. Added static gestures contributed to an extension of dynamic gesture recognition algorithm, proposed in ~\cite{hagrid}. There is the opportunity to recognize such dynamic gestures as ``zoom", ``click", and others.

\textbf{Content.} New samples were recorded by 28,394 unique crowdworkers, each located in their own scene. \cref{fig: chars}e-g shows the subjects' age, gender, and race distributions, calculated for all HaGRIDv2's 65,977 subjects. We preserve the HaGRID distribution and ensure domain-specific relevance by collecting samples in realistic indoor conditions with varying lighting conditions and subject-to-camera distances (see \cref{fig: hagrid_samples} in suppl. materials). The mean and standard deviation of HaGRIDv2 images' pixel values for the RGB channels are equal [0.54, 0.5, 0.474] and [0.234, 0.235, 0.231], respectively.

\textbf{Annotations.} HaGRIDv2 includes bounding box annotations for all hands. Each image has one or two boxes for one-handed gestures: one for the gesturing hand and one for the non-gesturing hand if it is in the frame (see \cref{fig: auto}a in the suppl. materials). Images with two-handed gestures strictly correspond to two boxes for each hand. Moving from the hand detection task to gesture detection, an additional box encompassing both hands is added to two-handed gesture images (see \cref{fig: auto}b-c in the suppl. materials). Bounding box annotations are proposed in COCO~\cite{coco} format with normalized relative coordinates.

\textbf{Splitting.} The dataset was divided into training (76\%), validation (9\%), and testing (15\%) sets, recorded by 57,656, 4,209, and 4,114 subjects, respectively. Note that training, validation, and test sets of original HaGRID are the subsets of corresponding HaGRIDv2 sets. \cref{fig: chars}a-c illustrate the subjects' distribution across three sets with improved heterogeneity in the test and validation sets compared to the training one. Sets are balanced in age, gender, brightness, and race due to randomness.

In addition, we provide hashed user IDs for researchers to split the dataset on their own. Also, such automatically received meta information as age, gender, and race for each subject, and keypoints for each hand on the images also supplied. Since the dataset is large, we also provided the lightweight version, with all images resized to 512 pixels on the shortest side.


\section{Dynamic Gesture Recognition Algorithm}
The original HaGRID~\cite{hagrid} authors proposed a dynamic gesture recognition algorithm that allows recognition based solely on static gestures, eliminating the need for video-based training. This paper presents a novel approach based on such a logic with extended functionality.

\begin{table*}[]
\centering
\scalebox{0.68}{
\begin{tabular}{lccccc}
\hline
\multirow{2}{*}{Model} & 
\multirow{2}{*}{Model size (MB)} & 
\multirow{2}{*}{Parameters (M)} & 
\multirow{2}{*}{Inference time (ms)} & 
\multicolumn{2}{c}{Metrics} \\ 
\cline{5-6} &  &  &  & F1-score & mAP \\
 \hline\hline
\multicolumn{6}{c}{Gesture Detection} \\ 
\hline
SSDLite MobileNetV3 Large~\cite{ssd} & 10.3 & 2.7 & 26.3 & - & 72.7 \\ 
YOLOv10n~\cite{yolov10} & 10.33 & 2.7 & 85.67 & - & 88.2 \\
YOLOv10x~\cite{yolov10} & 121.5 & 31.6 & 1145.6 & - & \textbf{89.4} \\ \hline\hline
\multicolumn{6}{c}{Full-Frame Classification} \\ 
\hline
ResNet18~\cite{resnet} & 42.7 & 11.2 & 37.8 & 98.3 & - \\
ResNet152~\cite{resnet} & 222.65 & 58.2 & 226.94 & \textbf{98.6} & - \\
MobileNetV3 Small~\cite{mobilenetv3} & 6 & 1.6 & 5.1 & 86.7 & - \\
MobileNetV3 Large~\cite{mobilenetv3} & 16.3 & 4.2 & 11.34 & 93.4 & - \\
ViTB16 (pretrained)~\cite{vit} & 327.4 & 85.8 & 350.9 & 91.7 & - \\
ConvNeXt Base~\cite{convnext} & 334.2 & 87.6 & 320 & 96.4 & - \\ \hline\hline
\multicolumn{6}{c}{Hand Detection} \\ 
\hline
YOLOv10n~\cite{yolov10} & 10.3 & 2.7 & 75.8 & - & 87.9 \\
YOLOv10x~\cite{yolov10} & 120.8 & 31.6 & 1150.7 & - & \textbf{88.8} \\ \hline
\end{tabular}}
\newline
\caption{Models training results on the HaGRIDv2. F1-score and Mean Average Precision (mAP) were chosen as classification and detection metrics, respectively. Intel(R) Xeon(R) Gold 6348 CPU @ 2.60GHz was used to compute inference time.}
\label{tabl:base_exps}
\end{table*}

\textbf{Algorithm.} \cref{fig: dyn_ges_scheme} in suppl. materials shows the pipeline of dynamic gesture recognition. Note that the algorithm processes each frame independently during inference without analyzing the entire sequence from start to end. We detect hands in each frame with a lightweight RFB~\cite{rfb_repo} model, ensuring faster inference. Further, the received crops are classified into gestures utilizing a single Residual Block~\cite{resblock}. To identify the specific gesture sequence boundaries accurately, we create a queue of the recognized crops over the last $n$ frames, where $n$ is experimentally set to 30. Besides, we implement checks on the duration of each gesture and track the location of its initiation and completion to ensure accurate classification. 

\textbf{New Dynamic Gestures.} The algorithm supports four categories of gestures: swipes, zooms, clicks, and drag-and-drops. Each category includes multiple gesture variations, as illustrated ~\cref{fig: dynamic_ges} in the suppl. materials. These variations allow for associating different functionalities with each gesture, enhancing the system's versatility.

Based on such an approach, the system is predictable and lightweight, with 276,292 parameters and 106.14 MFLOPs for the detector and 102,605 parameters and 6.9 MFLOPs for the classifier, and runs efficiently on standard CPU hardware. Additionally, it is easily expandable with new custom gestures, requiring only static images for training.

\section{Base Experiments}

\label{sec:base_exps}
\subsection{Experimental Setup}
The base experiments are divided into three groups: gesture detection, gesture classification, and hand detection. We resized the images' maximum side to 224 and padded the result to a square. The full-frame gesture classification is based on 33 main classes without the "no gesture" class, as each image contains one of the target gestures. For performance evaluation, we used the F1-score and Mean Average Precision (mAP) metrics for classification and detection tasks, respectively. All models were trained on a single Tesla H100 with 80GB for 100 epochs with a batch size of 128. Early stopping was triggered after 10 epochs without the metric increasing by at least 0.01. \cref{tabl:params} in the suppl. materials describes the detailed training hyperparameters.

\textbf{Hand and Gesture Detection.} Three detection architectures, SSDLite MobileNetV3 Large~\cite{ssd}, YOLOv10n, and YOLOv10x~\cite{yolov10}, were employed to ensure that HaGRIDv2 can train a robust gesture detector. We use the SGD optimizer with an initial learning rate of 0.01 for YOLOs and 0.0001 for SSDLite. The YOLO models employed default augmentations such as mosaic, hsv, and horizontal flips, while SSD trained without any modifications. Using the same setup, the YOLOv10x model was used to train for the hand detection task. 

\textbf{Gesture Classification.} In addition to ResNet-18, ResNet-152, MobileNetV3 Small, MobileNetV3 Large, and pre-trained on ImageNet ViTB16 utilized in ~\cite{hagrid}, we also employed ConvNext as a full-frame gesture classifier. The AdamW optimizer with a specified initial learning rate, weight decay, and scheduler for each architecture (see \cref{tabl:params} in the suppl. materials) was used.

\subsection{Results}
\cref{tabl:base_exps} presents the evaluation metrics on the HaGRIDv2 test subset. The metrics are remarkably high, demonstrating the dataset's effectiveness in training robust models. To ensure the model's ability to work in real-life conditions, we provide a demo of gesture classification and detection models in our repository. 

\section{Cross-Dataset Evaluation}
\textbf{Experimental Setup.} This section covers cross-dataset evaluation for hand and gesture detection\footnote{Cross-dataset evaluation for classification was excluded due to the lack of accessible datasets with sufficient overlapping gestures, making the comparison non-informative.}. We utilized the same setup across all experiments, employing YOLOv10n as a detector, the hyperparameters and augmentations described in \cref{sec:base_exps}, and mAP as a detection metric. 

\subsection{Gesture Detection}
\label{subsec:gesture_det_cross}
\textbf{Datasets.} We were limited to the HANDS and OUHANDS datasets in the gesture detection task since we could not access the LaRED and SHAPE datasets. These datasets intersect only in 5 classes — namely ``fist," ``one," ``palm," ``peace," and ``three" — with HaGRIDv2. Thus, we left only samples with these overlapping gestures, reducing three training, validation, and testing sets. The original HaGRID dataset was excluded from these experiments, as its content is entirely subsumed within HaGRIDv2.

\textbf{Results.} \cref{tabl:cross_gesture_det} indicates the HaGRIDv2's complexity, as evidenced by the lowest test average mAP. Furthermore, HaGRIDv2 exhibits superior domain generalization, supported by the highest train average mAP. Notably, HaGRIDv2 is the only dataset that consistently achieves valuable metrics across tests on other datasets, underscoring its value in training robust models.

\begin{table}[t]
\centering
\scalebox{0.7}{
\begin{tabular}{l||ccc||c}
\hline
Trained / Tested & \cite{ouhands} & \cite{hands} & HaGRIDv2 & Train avg. mAP (\textuparrow) \\ \hline\hline
OUHANDS~\cite{ouhands} & \cellcolor{LightCyan}65 & 0.07  & 4.25 & 23.1 \\ 
HANDS~\cite{hands} & 0.0 & \cellcolor{LightCyan}65.4  & \textbf{7} & 24.1 \\ 
HaGRIDv2 (ours) & \textbf{67} & \textbf{66.3} & \cellcolor{LightCyan}88.1 & \textbf{73.8}           \\ \hline\hline
Test avg. mAP (\textdownarrow) & 44 & 43.9  & \textbf{33.1} & \\ \hline
\end{tabular}}
\caption{Cross-dataset evaluation in the gesture detection task. mAP was computed for each pair of datasets and averaged separately for training and testing. Higher average mAP during training indicates greater model robustness, while lower mAP during testing reflects higher dataset complexity. Diagonal values were excluded from the averages to ensure unbiased comparison and assess generalization.}
\label{tabl:cross_gesture_det}
\end{table}

\begin{table*}[]
\centering
\scalebox{0.67}{
\begin{tabular}{l||ccccc||c}
\hline
Trained / Tested & EgoHands & BodyHands & Human-Parts & HaGRID & HaGRIDv2 & Train avg. mAP (\textuparrow) \\ \hline\hline
EgoHands~\cite{egohands} & \cellcolor{LightCyan}75.4 & 0.9 & 1.16 & 3.54 & 4.15 & 2.4 \\ 
BodyHands~\cite{bodyhands} & 14 & \cellcolor{LightCyan}35.2 & 26.4 & 49.4 & 37.3 & 31.7 \\ 
Human-Parts~\cite{human_parts} & 35.7 & \textbf{21.8} & \cellcolor{LightCyan}50.5 & 58.7 & 53.3 & 42.4\\ 
HaGRID~\cite{hagrid} & 39.2 & 19.7 & \textbf{35} & \cellcolor{LightCyan}86.7 & \textbf{73.9} & 42 \\
HaGRIDv2 (ours) & \textbf{39.4} & 17.3 & 34.9 & \textbf{87.2} & \cellcolor{LightCyan}87.9 & \textbf{44.7} \\ \hline\hline
Test avg. mAP (\textdownarrow) & 32 & \textbf{14.9} & 24.4 & 49.7 & 42.2 & \\ \hline
\end{tabular}}
\caption{Similar cross-dataset evaluation as in \cref{tabl:cross_gesture_det} for hand detection task.}
\label{tabl:cross_hand_det}
\end{table*}

\subsection{Hand Detection}
\textbf{Datasets.} We compare HaGRIDv2 with datasets designed explicitly for hand detection to assess its ability to solve this task. Since HaGRIDv2 was annotated with vertical bounding boxes, we only compared it with similarly annotated BodyHands, Human-Parts, and EgoHands datasets. We also included the original HaGRID dataset to test the impact of more quantity and variety in gestures in the HaGRIDv2. The training, validation, and test sets of HaGRID and HaGRIDv2 were carefully curated to prevent overlap, eliminating the risk of data leakage and ensuring the integrity of the comparison.

\textbf{Results.} Although neither version of HaGRID was initially designed for hand detection, and their samples appear simpler than other datasets (as seen in the \cref{fig: samples} in the suppl. materials), \cref{tabl:cross_hand_det} demonstrates that HaGRIDv2 improves the generalization ability on hand detection task. Additionally, HaGRIDv2 achieves a higher mAP on HaGRIDv2 than HaGRID itself, suggesting that increasing the diversity of gesture classes improves the model's ability to generalize older gestures.

\section{HaGRID vs HaGRIDv2}
\subsection{Pre-train Impact}
\textbf{Experimental Setup.} We compared HaGRID and HaGRIDv2 as pre-training datasets to demonstrate that a 2$\times$ increase in samples and class diversity in HaGRIDv2 consistently yields more reasonable results. ResNet18 for gesture classification and YOLOv10 for gesture detection were employed, applying the same hyperparameters and metrics from base experiments in \cref{sec:base_exps}.

\textbf{Datasets.} HANDS and OUHANDS were utilized to fine-tune detectors on their training sets with further assessment on their test sets. The pre-trained classifiers were fine-tuned on the Kinect Leap, Senz3D, OUHANDS, and HANDS training sets, while HTU HGR and SIT-HANDS were not available due to the invalid link.

\begin{figure}[t]
  \centering
  \includegraphics[height=0.6\linewidth]{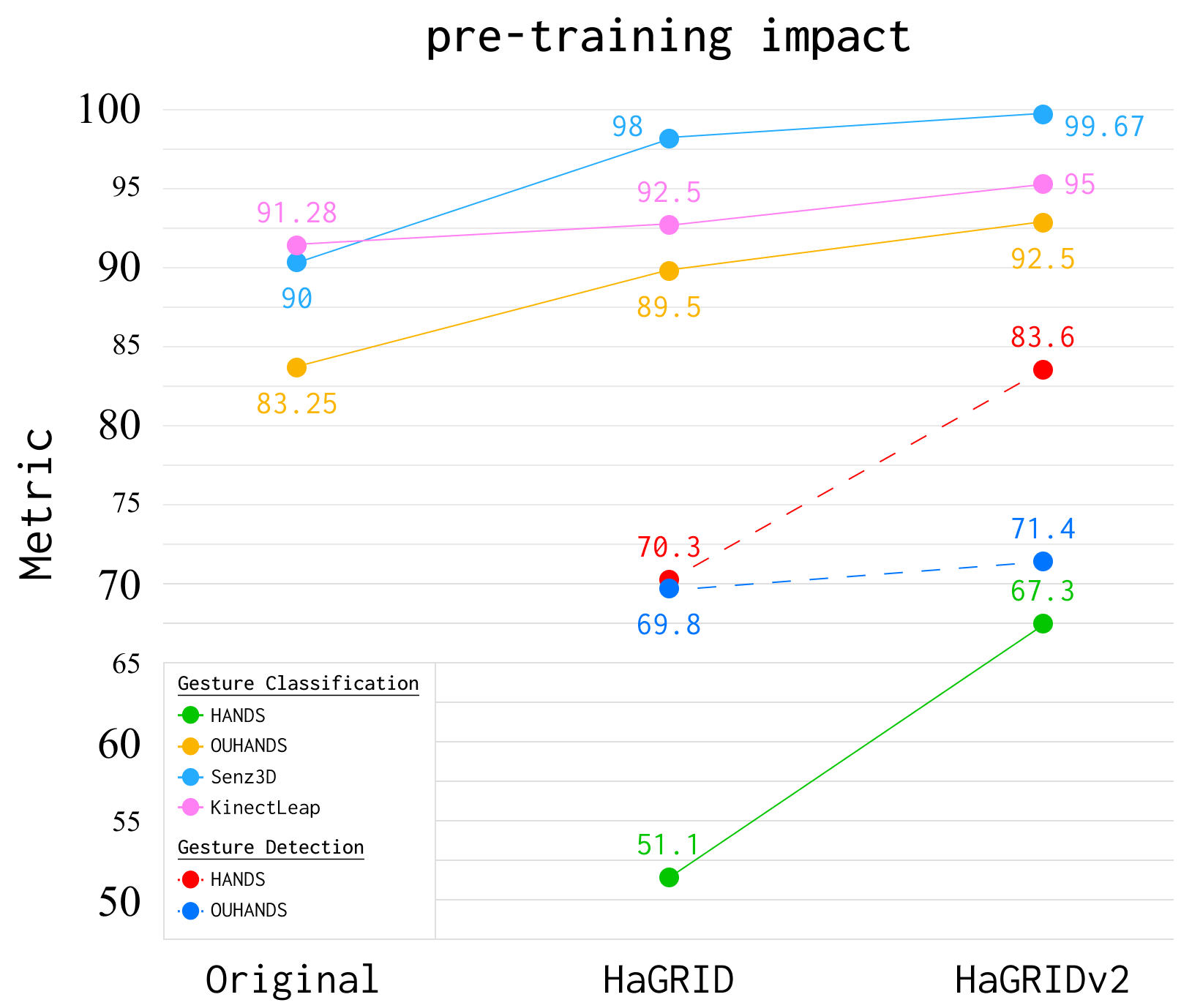}
  \caption{Impact of pre-training on gesture classification and detection across HaGRID and HaGRIDv2. ``Original" metrics are sourced from the respective dataset papers; missing values indicate metrics not reported by the authors.}
  \label{fig: pretrains_classification}
\end{figure}

\textbf{Results.} \cref{fig: pretrains_classification} shows that models pre-trained on HaGRIDv2 consistently outperformed those pre-trained on HaGRID, improving model generalization and proving its indispensability for pre-training.

\begin{figure}[t]
  \centering
  \includegraphics[width=0.75\linewidth]{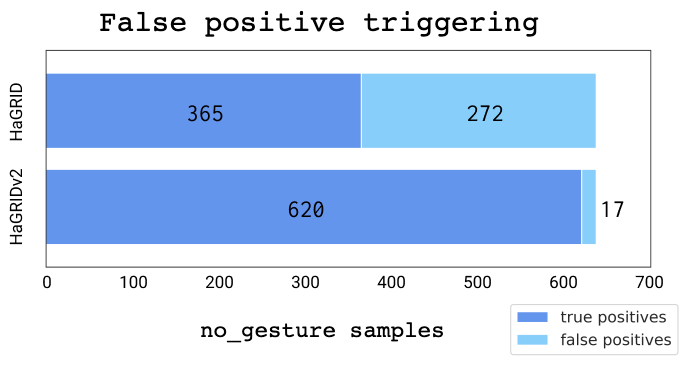}
  \caption{Comparing the false positives on the ``no gesture" class for HaGRID and HaGRIDv2 datasets.}
  \label{fig: false}
\end{figure}

\subsection{False Positive Triggering}
\textbf{Experimental Setup.} We conducted a check for false positive triggering to assess the impact of diversifying the ``no gesture" class (see \cref{fig: no_gesture_samples} in suppl. materials). We trained two YOLOv10n detectors: one on the original HaGRID dataset and the other on HaGRIDv2. Each trained model was assessed on the ``no gesture" samples from the HaGRIDv2 test set to evaluate the amount of false positives.

\textbf{Results.} The HaGRIDv2-trained model produces 6 times fewer false positive errors than the HaGRID-trained model, which is especially important for production-level HGR systems (see \cref{fig: false}). The mAP metrics were 57 for HaGRID and 72.9 for HaGRIDv2, highlighting the superiority of the new ``no gesture" configuration. 

\subsection{Gesture Generation}
\label{subsec: gest_gen}
\textbf{Experimental Setup.} This section aims to demonstrate that adding new gestures improves the quality of generating images of people showing gestures. We fine-tuned two Stable Diffusion 2.1 models using the LoRA~\cite{lora} (Low-Rank Adaptation) from the diffusers library~\cite{diffusers} on the HaGRID and HaGRIDv2 datasets. Each image in the datasets was annotated with an automatically generated description using BLIP-2~\cite{blip}, following the format: \textit{``\{blip\_caption\} showing \{gesture\_name\} gesture".} During sample generation, we used prompts like \textit{``there is a person showing \{gesture\_name\} gesture"}. To conduct a fair comparison, we also compared the HaGRIDv2 fine-tuned model with the original Stable DIffusion 2.1 without fine-tuning.

\textbf{Evaluation.} To compare HaGRID and HaGRIDv2, we evaluated the generation quality of 18 gestures from the original HaGRID. For the original diffusion model, we tested 26 gestures from HaGRIDv2, excluding ``inverted" and ``thumb\_index" gestures due to their specificity, and kept only one version of each gesture (e.g., keeping ``heart" and removing ``heart2"). Each model generated three images per gesture, totaling 6 images per gesture. We conducted a Subjective Benchmark Scoring (SBS), as shown in \cref{fig: sbs} in the suppl. materials. Three crowdworkers compared each pair of images, resulting in 162 comparisons for the fine-tuned models and 234 for the original Stable Diffusion and HaGRIDv2-tuned models. Workers voted on two criteria: (1) which image had more anatomically accurate hands and (2) which better resembled the reference gesture.

\textbf{Results.} \cref{fig: sbs2} in the suppl. materials shows that the model fine-tuned on HaGRIDv2 generates more accurate gestures due to the wider variety of classes it was trained on, allowing for easier replication of specific patterns. However, the original HaGRID achieved slightly better anatomical accuracy due to its simpler distribution and lack of complex hand postures. Additionally, the comparison with the original Stable Diffusion 2.1 model indicated its limited ability to generate recognizable gestures, with the anatomical accuracy of the hands also being inferior (see \cref{fig: gen_gest} in the suppl. materials).

\section{Ethical Consideration}
\label{sec: ethic}
\textbf{Dataset Creation.} HaGRIDv2's samples contain personal information, so crowdworkers must consent to collect, process, and publish their photos. We comply with Russia's Federal Law "On Personal Data" (27.07.2006 N152), ensuring legal data handling. For ethical reasons, images of children were excluded from the dataset. Crowdworkers involved in all stages of data collection were compensated at least the minimum wage in Russia. After validation, we justified each rejected photo and allowed crowdworkers to challenge rejections. As the HaGRID dataset is part of HaGRIDv2, we ensured that HaGRID adheres to the described ethical requirements by contacting its authors.

\textbf{Biases.} Utilizing only Russian crowdsourcing platforms can lead to an imbalance in the racial diversity of workers. We tried to minimize this gap and covered the most frequently identified races -- Caucasian, Negroid, and Mongoloid. Even though there is an imbalance, trained on the HaGRIDv2 neural network can accurately recognize gestures from underrepresented racial groups.

\textbf{Possible Misuse.} There is the risk of misusing the datasets with faces to improve surveillance systems, profile individuals based on race, create deepfakes, and contribute to identity theft. We release the dataset under a public license for non-commercial use in research purposes, acknowledging the potential risk of its misuse for unlawful activities.
We used anonymized user hash IDs in the dataset annotation to preserve crowdworkers privacy and enable the ability to split HaGRIDv2 by its users.

\section{Limitations}
\label{sec: limitations}
\textbf{Dataset.} As noted in \cref{sec: ethic}, the dataset is biased towards the white race, potentially compromising the algorithm's robustness and performance in diverse scenarios. Additionally, the Gaussian age distribution, limited to individuals 18 years and older, may lead to inaccurate predictions for children and seniors. The ``no gesture" class expansion is specified to poses typical for device interactions, potentially leading to false positives in other scenarios. Furthermore, predominantly home-based scenes may lead to incorrect operation when people wear outdoor clothing and gloves in different weather conditions.

\textbf{Dynamic Gesture Recognition Algorithm}. The deterministic nature of the algorithm reduces its robustness, as it demands users to perform gestures with exact precision. This inflexibility can result in recognition inconsistencies, especially in real-world scenarios where slight variations in gesture execution are common, ultimately limiting the algorithm's reliability and user experience.

\textbf{Generative Models.} While the HaGRIDv2 dataset aids in training models to generate people displaying gestures (see \cref{subsec: gest_gen}), it has limitations. The dataset's focus on specific gestures and the limited variety of hands in natural positions restrict the models' ability to generate anatomically accurate hands in free poses, which may limit its broader applicability. Additionally, since the model was fine-tuned solely on samples from HaGRIDv2's limited distribution, it tends to generate images that look quite similar, reducing the overall variety.

\section{Conclusion}
\label{sec: conclusion}
This paper introduces HaGRIDv2, an enhanced version of HaGRID, which become the largest and most diverse gesture dataset for HGR systems. Including a new ``no gesture" class significantly strengthens the system's robustness and adaptability to real-world conditions, paving the way for more reliable gesture-based interaction technologies. The HaGRIDv2 can also be used for robust hand generation by diffusion models. We also present a dynamic gesture recognition algorithm that identifies various manipulative gestures for device control. The dataset, pre-trained models, and the dynamic gesture recognition algorithm will be published in our repository.

{\small
\bibliographystyle{ieee_fullname}
\bibliography{egbib}
}
\clearpage
\onecolumn
\appendix
\section*{Supplementary materials}

\centering
\begin{figure*}[htp]
  \centering
  \includegraphics[width=0.8\linewidth]{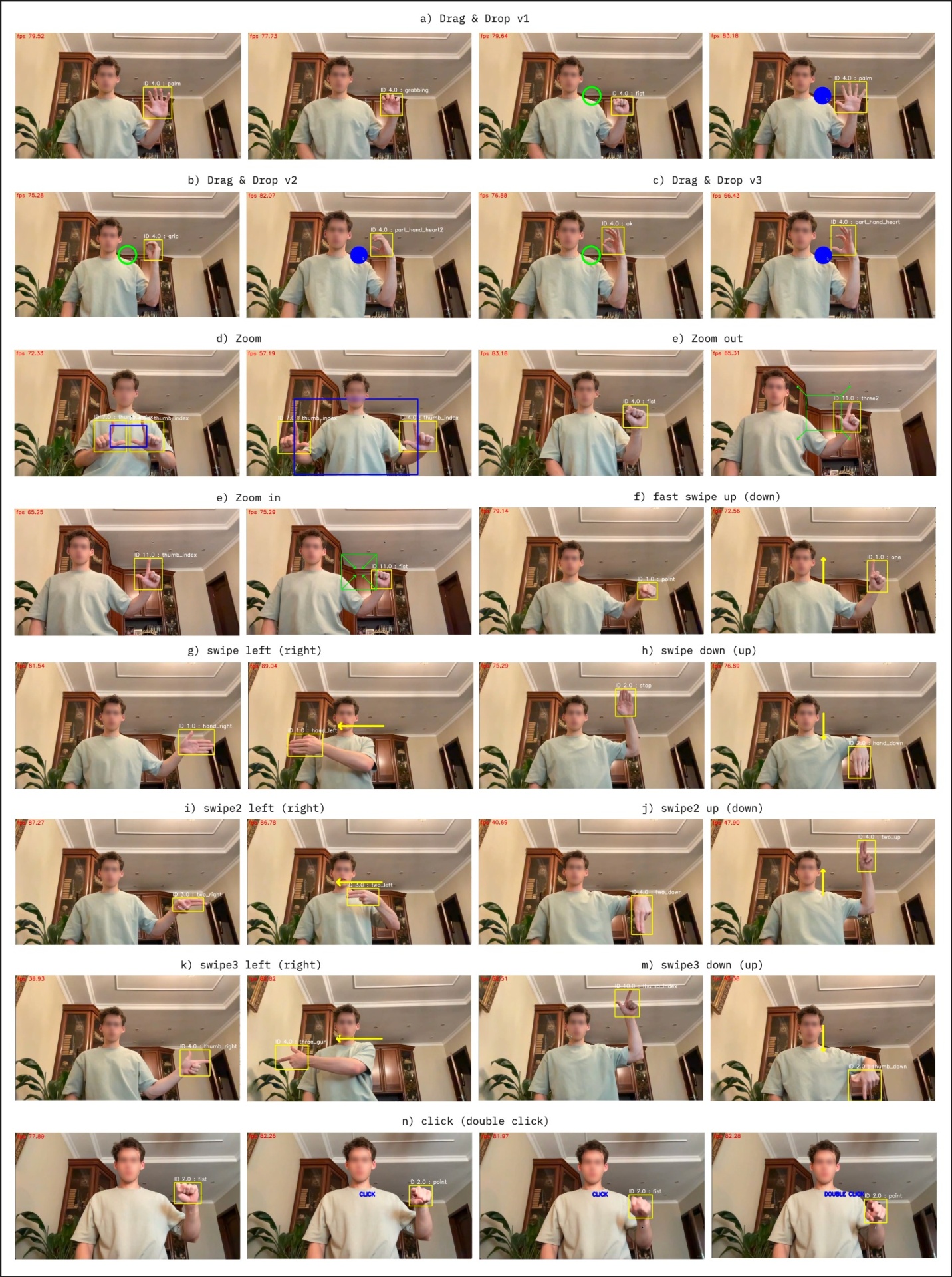}
  \caption{The screenshots from the dynamic gesture recognition demo. The bounding boxes highlight detected gestures with their class labels. Each dynamic gesture is marked related to its function: yellow arrows indicate swipe directions, green and blue circles represent drag and drop, respectively, ``click" and ``double-click" display their corresponding gestures, green arrows or a stretchable blue rectangle for zoom gestures.}
  \label{fig: dynamic_ges}
\end{figure*}

\begin{figure*}[htp]
  \centering
  \includegraphics[width=0.93\linewidth]{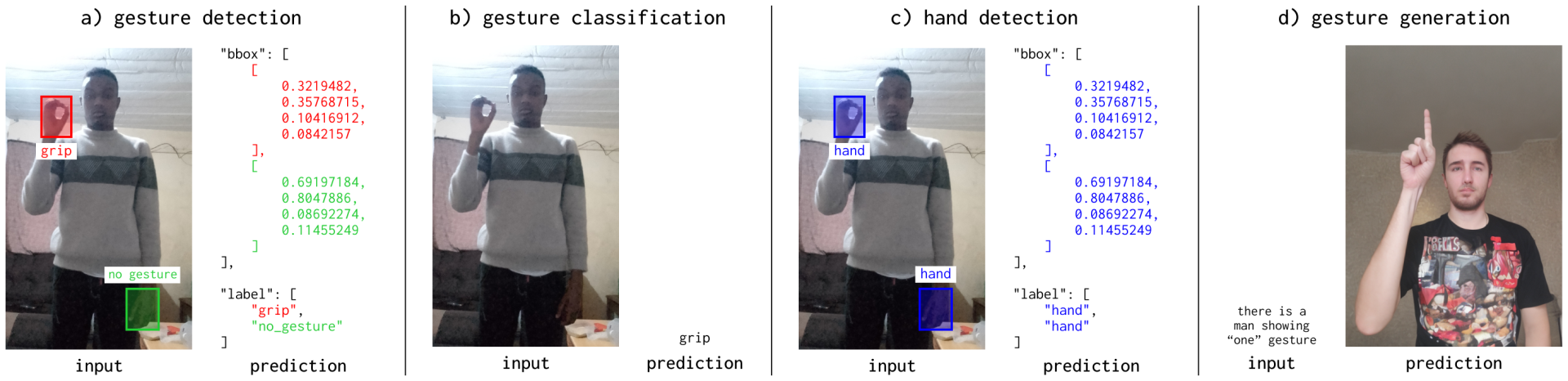}
  \caption{Different tasks addressed by HaGRIDv2. (a) The gesture detector aims to predict a bounding box with a label for each hand on the image; (b) The gesture classifier produces a label for the entire image; (c) The hand detector recognizes all hands by bounding boxes with the same label ``hand"; (d) The gesture generator creates an image of a person showing a gesture according to the prompt.}
  \label{fig: tasks}
\end{figure*}

\begin{figure*}[htp]
  \centering
  \includegraphics[width=0.5\linewidth]{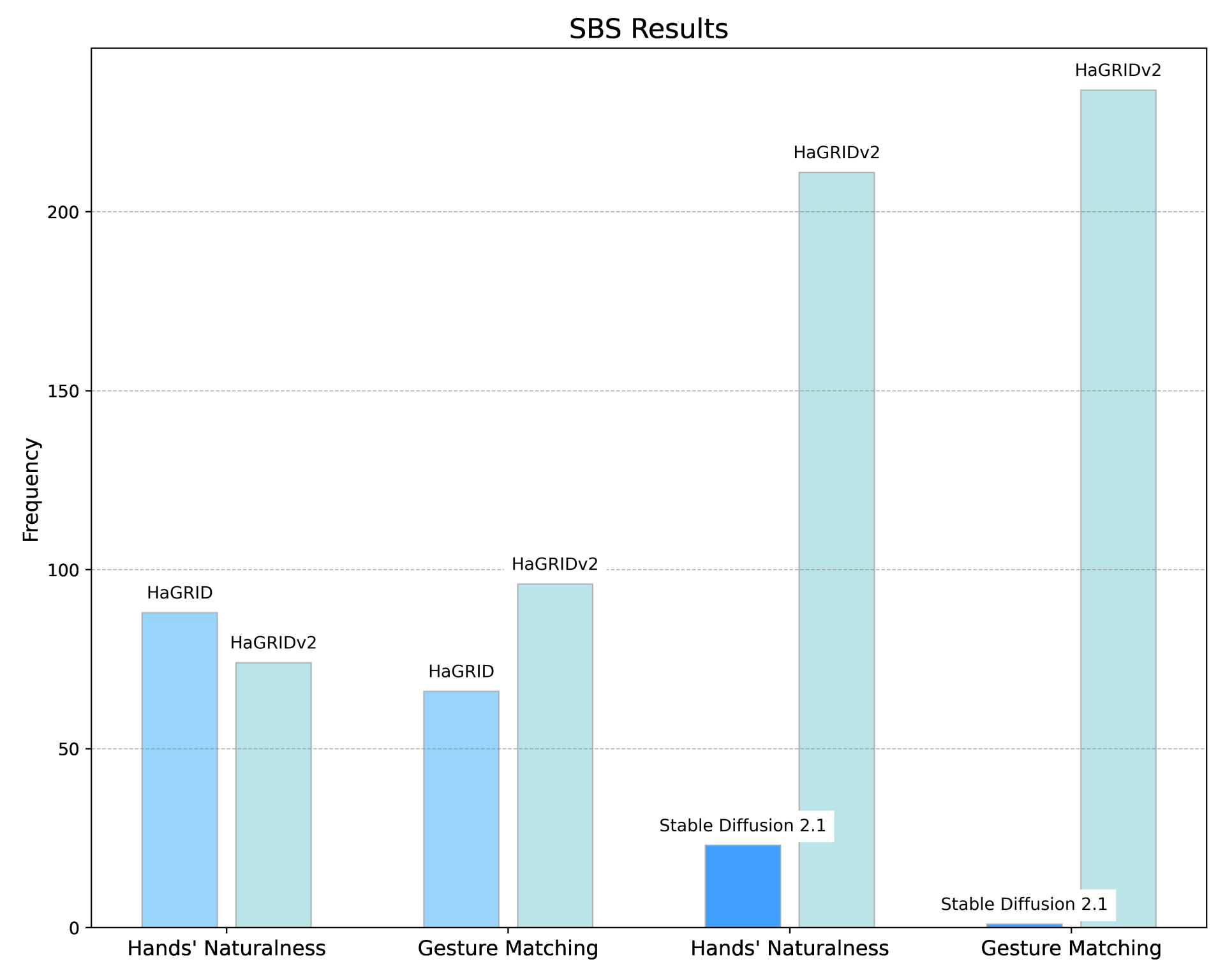}
  \caption{The SBS results compare Stable Diffusion 2.1 fine-tuned on the HaGRID and HaGRIDv2 datasets, as well as a comparison between HaGRIDv2 and the original Stable Diffusion 2.1 model.}
  \label{fig: sbs2}
\end{figure*}

\begin{figure*}[htp]
  \centering
  \includegraphics[width=0.7\linewidth]{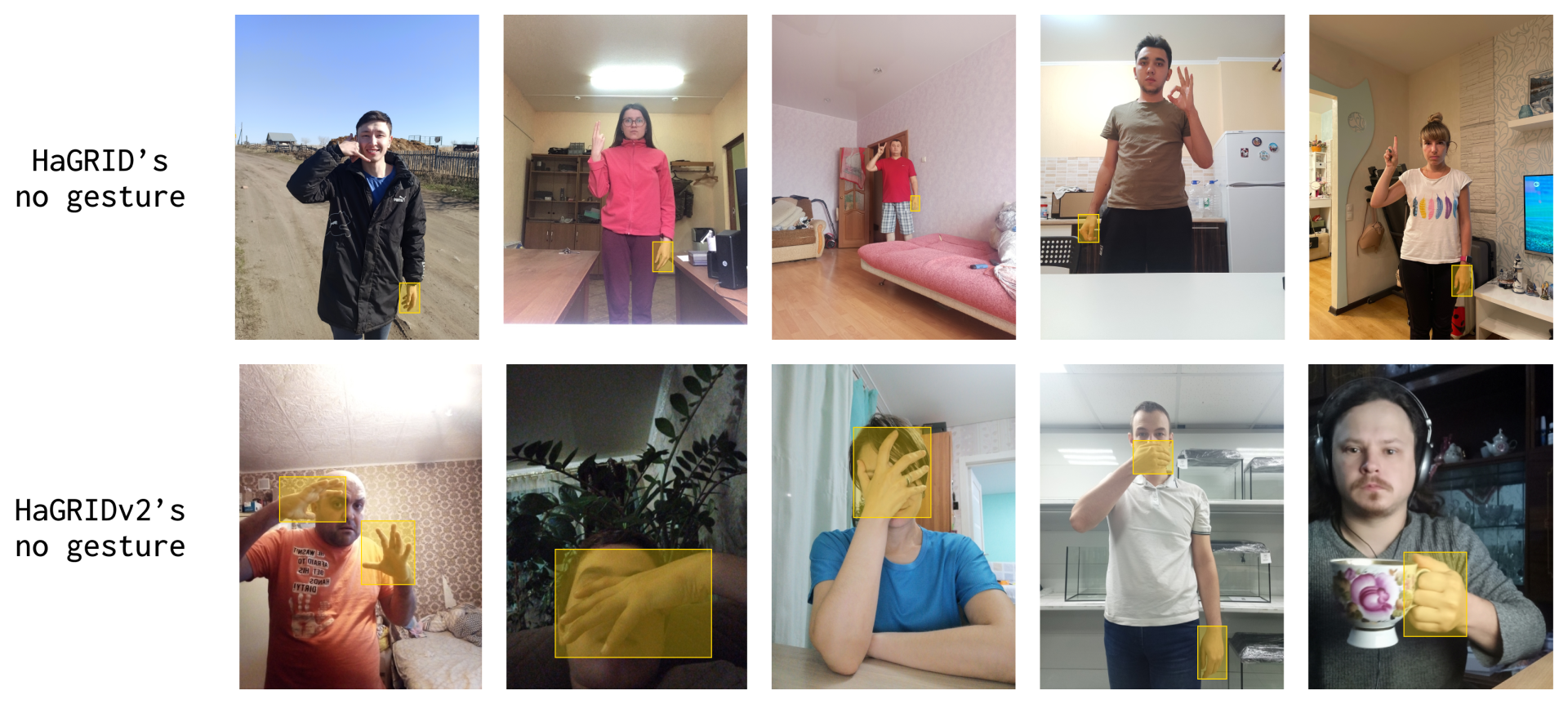}
  \caption{Samples of the ``no gesture" class in HaGRID and HaGRIDv2 datasets.}
  \label{fig: no_gesture_samples}
\end{figure*}

\begin{figure*}[htpb] 
  \centering
  \includegraphics[width=0.8\linewidth]{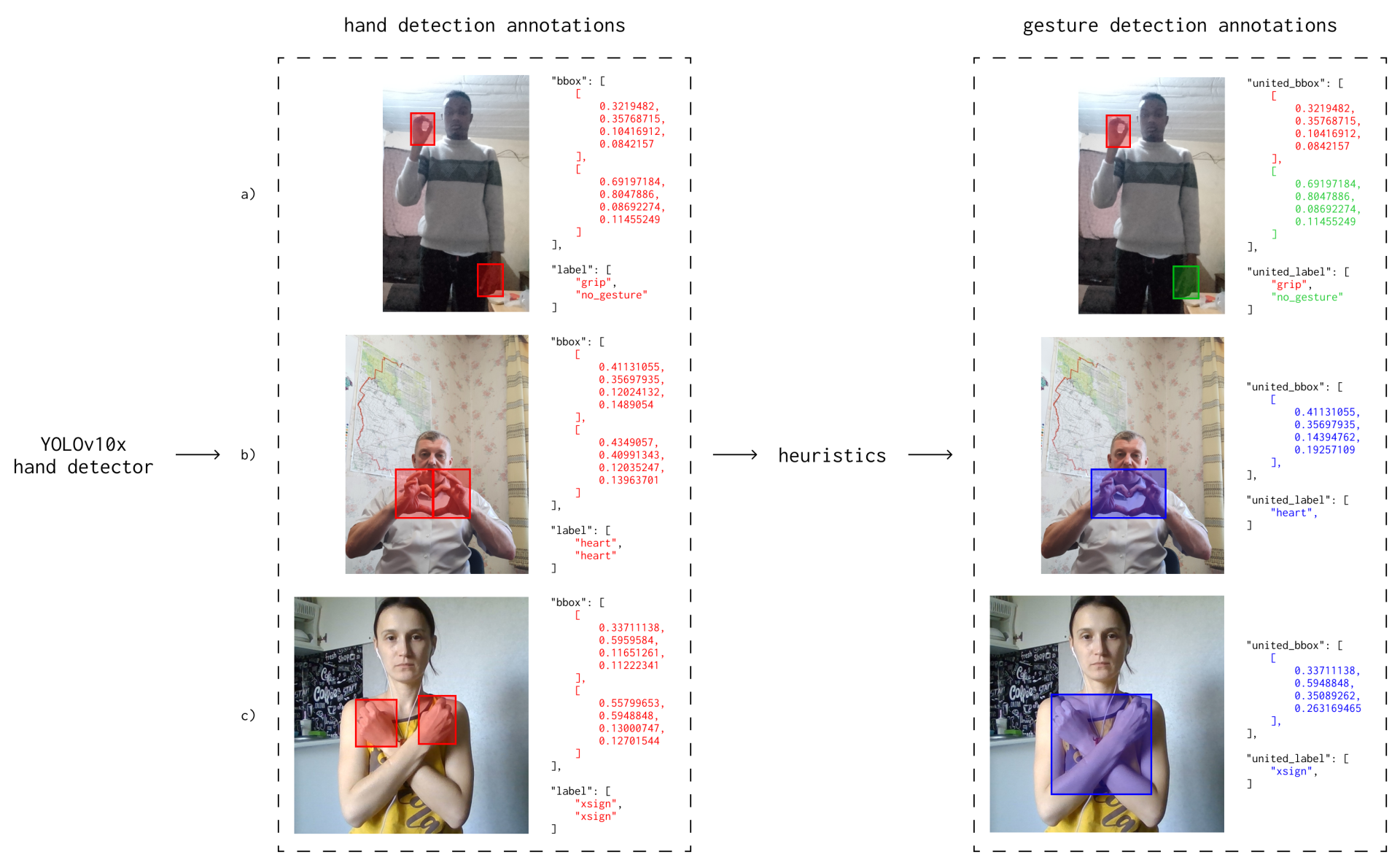}
  \caption{The pipeline for automatic image annotation. a) The higher hand on the one-handed gesture image is marked as the gesticulating hand; b) Two predicted boxes for the two-handed gesture are merged into a single box. c) A two-handed ``xsign" gesture is marked by a square bounding box, received by stretching a vertical line equal to the distance between two boxes.}
  \label{fig: auto}
\end{figure*}

\begin{table}
\begin{center}
\scalebox{0.75}{
\begin{tabular}{|p{1.0in}|p{2.3in}|p{1.0in}|p{2.3in}|}
\hline
Gesture & Applications & Gesture & Applications\\
\hline
thumb\_index & 
-- Input number 2 \newline
-- Mouse control on the screen \newline
-- Used for ZOOM and SWIPE dynamic gestures 
& point & 
-- Mouse control \newline
-- Activate smart home system by pointing \newline
-- Used for CLICK / DOUBLE CLICK / SWIPE dynamic gestures \\
\hline

thumb\_index2 & 
-- Screenshot of a specific area \newline
-- Start screen recording / sharing \newline
-- Take a selfie / screenshot \newline
-- Used for ZOOM dynamic gesture 
& pinkie & 
-- Input number 1 \newline
-- Set volume / brightness to minimum \\
\hline

middle\_finger & 
-- Express disapproval during a video conference \newline
-- Negative content rating 
& holy & 
-- Express a request during video conferences \newline
-- Switch to silent mode \newline
-- Play relaxing music or personal playlist \\
\hline

grip & 
-- Input number 0 \newline
-- Mute \newline
-- Give a negative (zero) rating \newline
-- Used for Drag-and-Drop dynamic gesture 
& grabbing & 
-- Move objects on the screen \newline
-- Used for Drag-and-Drop dynamic gesture \\
\hline

three3 & 
-- Input number 3 \newline
-- A humorous gesture to use during a video conference. It can be accompanied by a mask or stickers 
& timeout & 
-- Pause content \newline
-- Power off the system \newline
-- Emotional gesture during conversation (request to stop speaking). Mute the interlocutor / turn off the volume \\
\hline

take\_photo & 
-- Take a photo / screenshot / selfie \newline
-- Open a new window \newline
-- Start screen recording / sharing 
& xsign & 
-- Shut down the entire system \newline
-- Pause content \newline
-- Mute \newline
-- Report inappropriate content \\
\hline

three\_gun & 
-- Humorous gestures during video conferences, accompanied by stickers, fonts, or music \newline
-- Set the volume to maximum \newline
-- Turn on the music 
& heart & 
-- Express love during a video conference, accompanied by stickers, fonts, or music \newline
-- Like a song / video / add to playlist \\
\hline

\end{tabular}}
\end{center}
\caption{Gestures applications in gesture recognition systems. ``heart2" was not included as it has the same meaning as ``heart".}
\label{tabl:funcs}
\end{table}

\begin{figure*}[htp]
  \centering
  \includegraphics[width=0.8\linewidth]{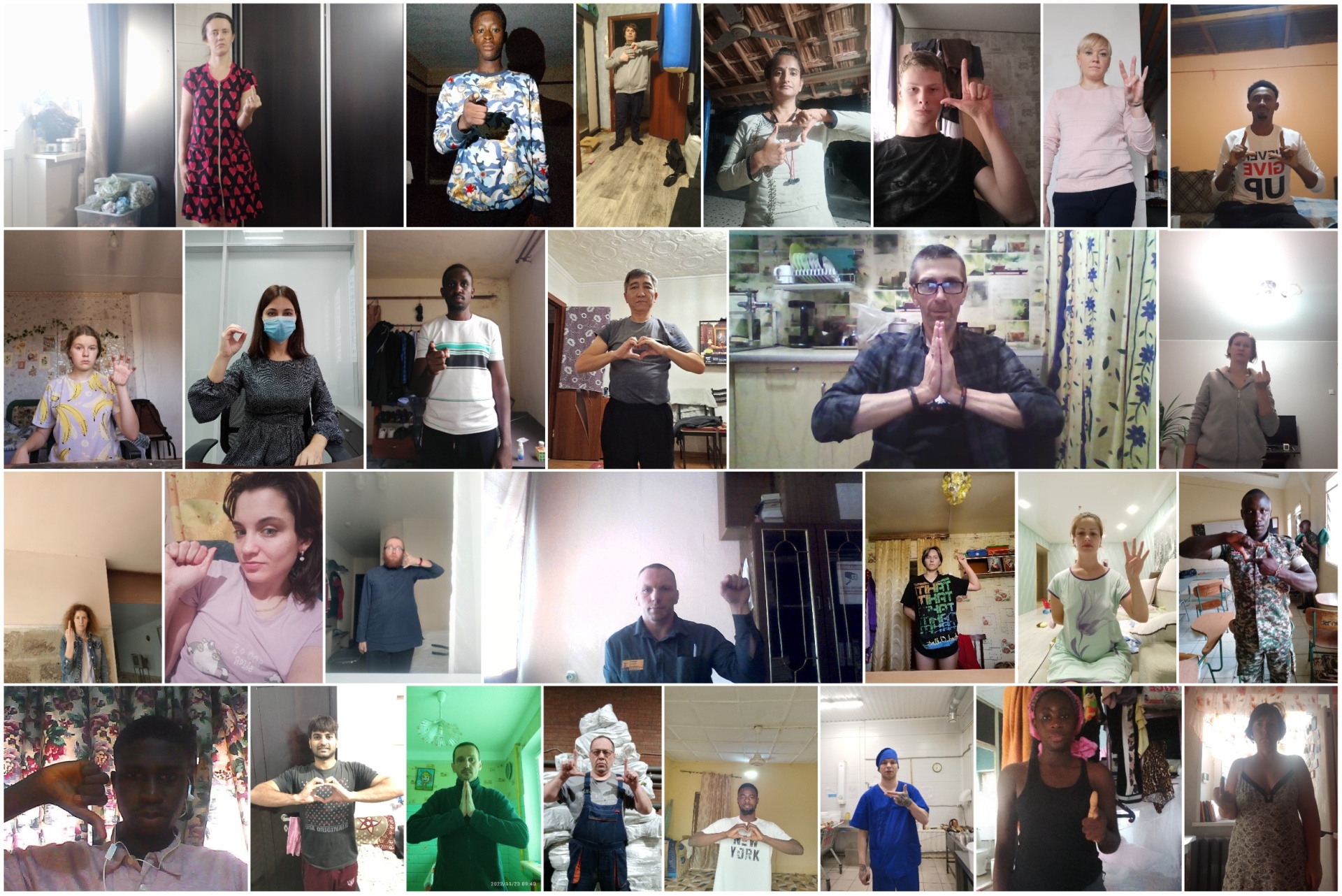}
  \caption{Samples from HaGRIDv2 dataset.}
  \label{fig: hagrid_samples}
\end{figure*}

\begin{figure*}[htp]
  \centering
  \includegraphics[width=1.0\linewidth]{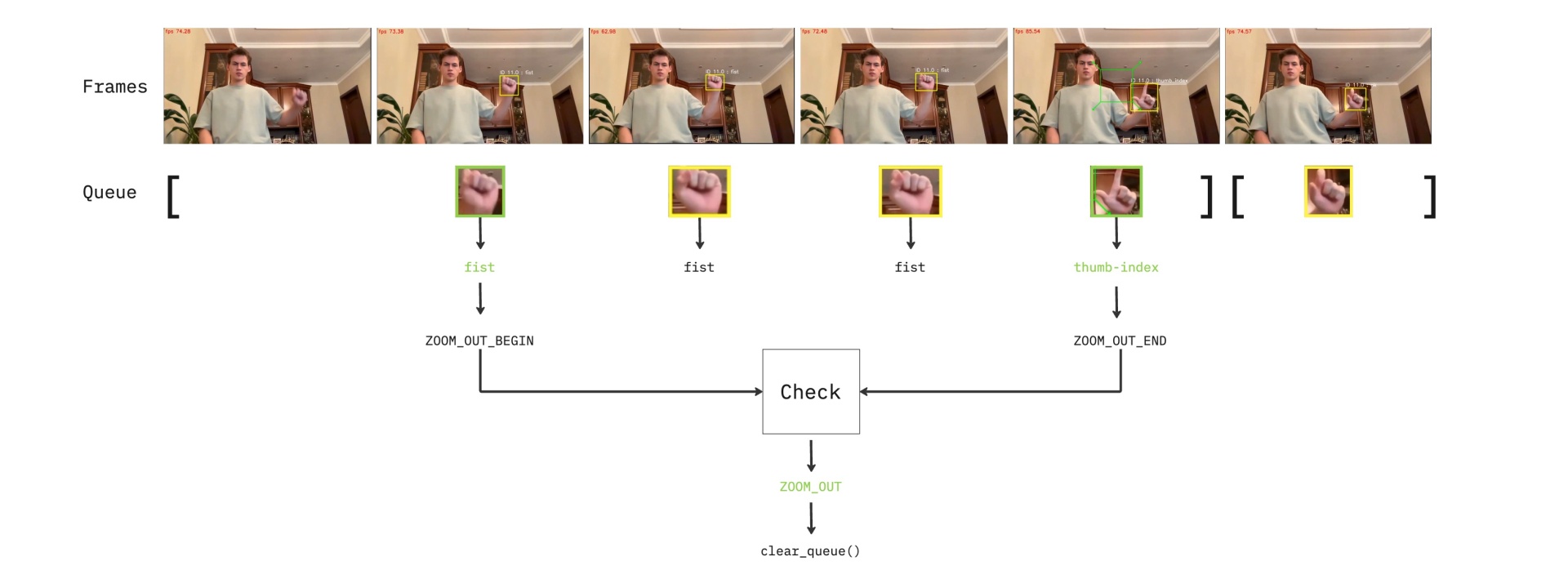}
  \caption{The algorithm for recognizing dynamic gestures, exemplified by the ``zoom out" gesture.}
  \label{fig: dyn_ges_scheme}
\end{figure*}

\begin{table*}[htp]
\centering
\scalebox{0.7}{
\begin{tabular}{lccccc}
\hline
Model & Optimizer & Weight Decay & Learning Rate & Scheduler & Scheduler' Params.\\
\hline
ResNet & SGD & $1^{-4}$ & $1^{-1}$ & ReduceLROnPlateau & mode: min, factor: 0.1\\
MobileNetV3 & SGD & $5^{-4}$ & $5^{-3}$ & StepLR & step size: 30, gamma: 0.1\\
VitB16 & SGD & $5^{-4}$ & $5^{-3}$ & CosineAnnealingLR & T max: 8\\
ConvNext & AdamW & $5^{-2}$ & $4^{-3}$ & CosineAnnealingLR, LinearLR & T max: 8, factor: 0.001\\
SSDLite & SGD & $5^{-4}$ & $1^{-4}$ & StepLR & step size: 30, gamma: 0.1\\
YOLOv10 & SGD & $5^{-4}$ & $1^{-2}$ & LambdaLR & sinusoidal function\\
\hline
\end{tabular}}
\caption{Training hyperparameters.}
\label{tabl:params}
\end{table*}

\begin{figure*}[htp]
  \centering
  \includegraphics[width=0.8\linewidth]{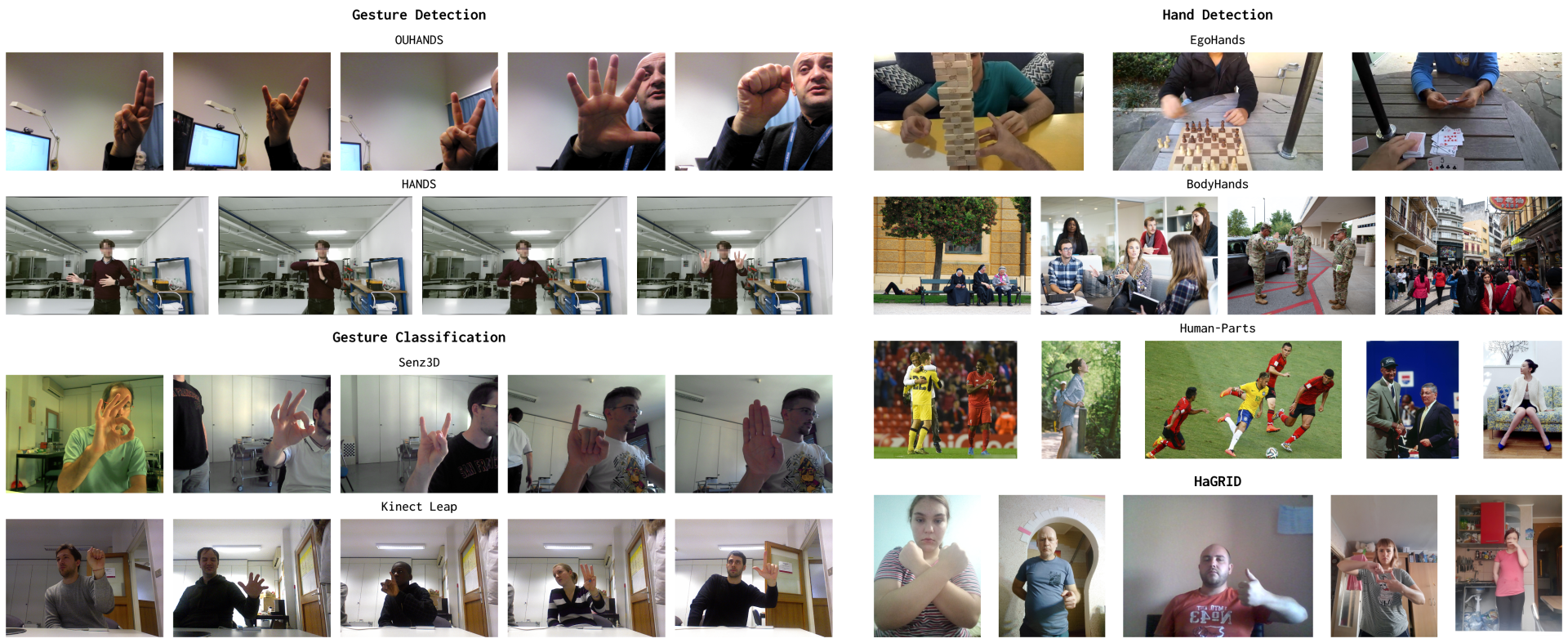}
  \caption{Samples from datasets, participated in cross-dataset evaluation and pre-trains impact experiment.}
  \label{fig: samples}
\end{figure*}

\begin{figure*}[htp]
  \centering
  \includegraphics[width=0.55\linewidth]{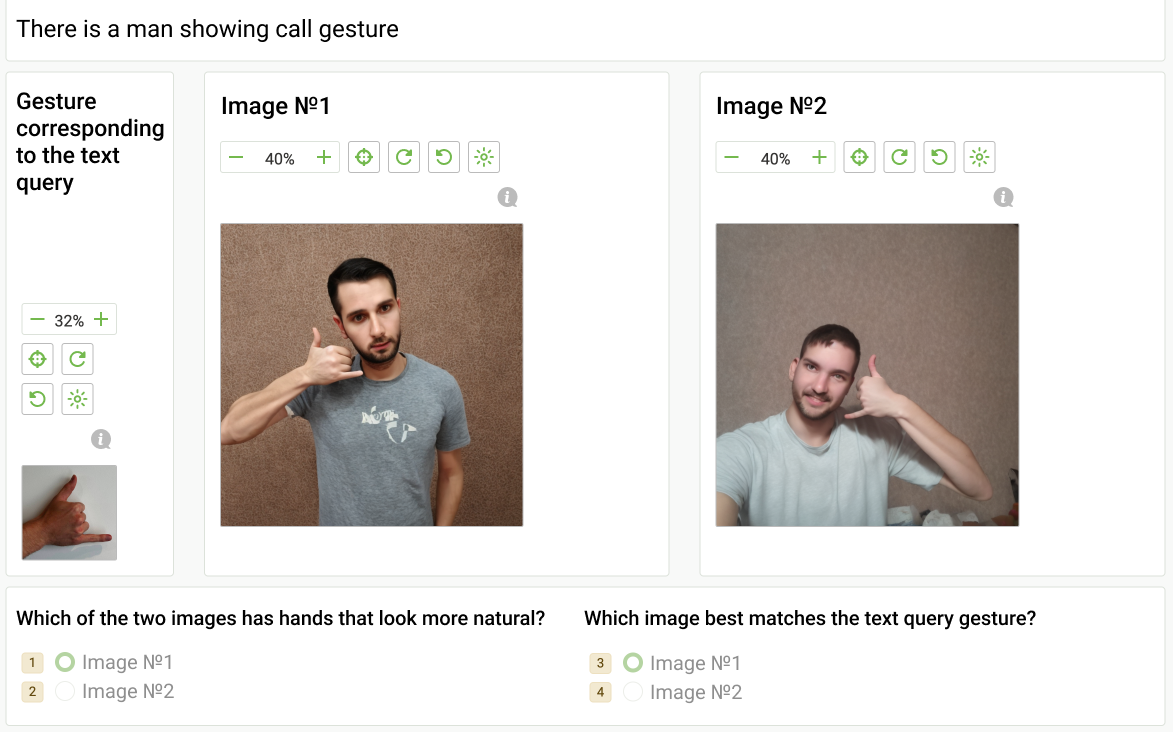}
  \caption{The interface for SBS evaluation presenting two images generated by two Stable Diffusion 2.1 models fine-tuned on HaGRID and HaGRIDv2 datasets. Crowdworkers are asked to assess these images based on two criteria: which hand posture appears more natural and which image better corresponds to the text query describing the gesture.}
  \label{fig: sbs}
\end{figure*}

\begin{figure*}[htp]
  \centering
  \includegraphics[width=0.65\linewidth]{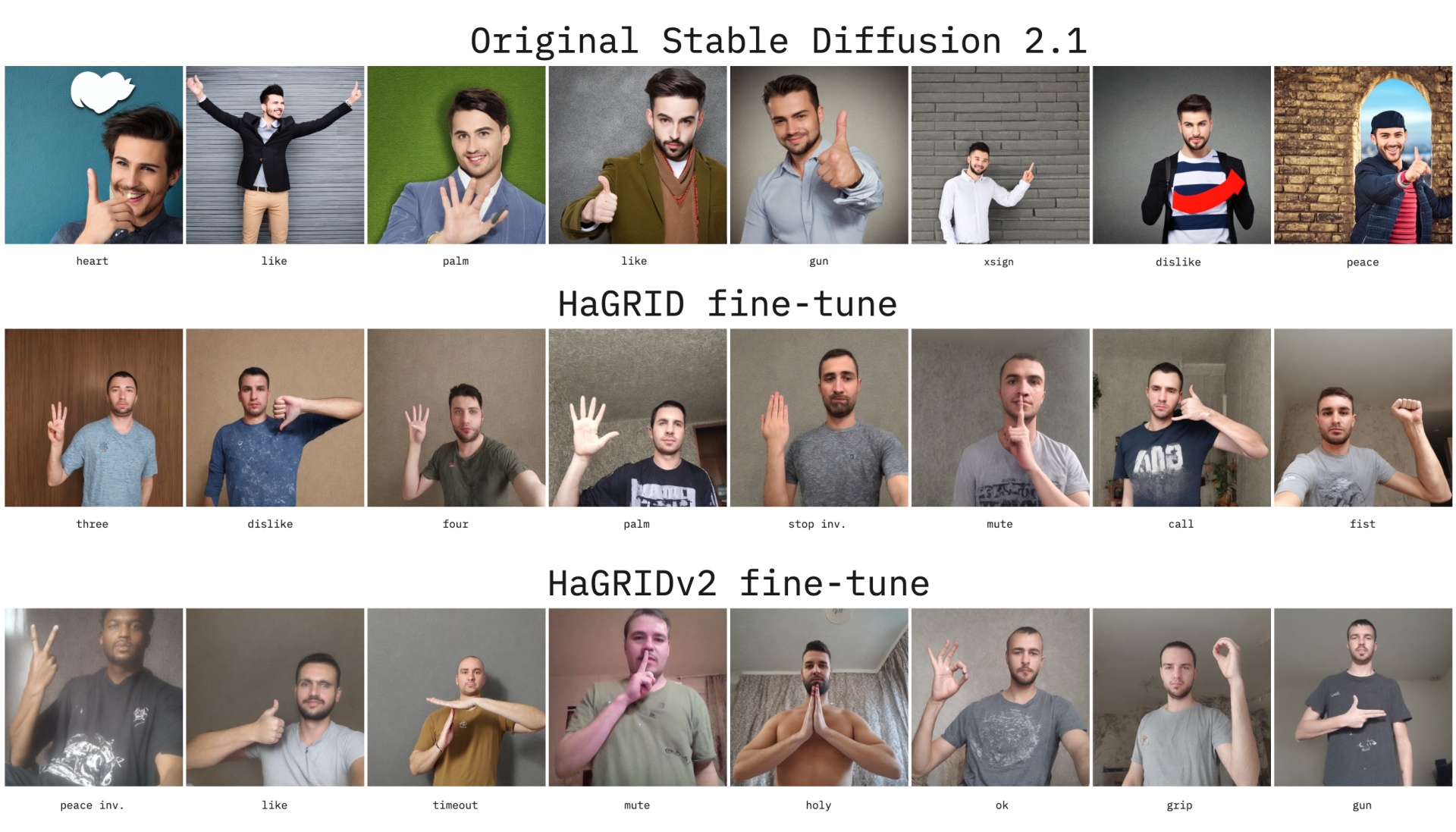}
  \caption{Examples of generated images with three Stable Diffusion 2.1 models: original, fine-tuned on HaGRID, and fine-tuned on HaGRIDv2.}
  \label{fig: gen_gest}
\end{figure*}

\end{document}